\documentclass[conference,a4paper]{IEEEtran}

\usepackage[utf8]{inputenc}

\usepackage[english]{babel}

\usepackage{graphicx}

\usepackage{multirow,booktabs}

\usepackage{amsmath,amssymb,amsfonts,amsthm}
\usepackage{nicefrac}
\usepackage[autolanguage]{numprint}
\usepackage{algorithm,algorithmic}

\usepackage{notations}

\usepackage[obeyDraft]{todonotes}
\usepackage[draft]{hyperref} 

\title{Predictive support recovery with TV-Elastic Net penalty and logistic regression: an application to structural MRI}
\author{%
\IEEEauthorblockN{%
 Mathieu Dubois\IEEEauthorrefmark{1},
 Fouad Hadj-Selem\IEEEauthorrefmark{1},
 Tommy Löfstedt\IEEEauthorrefmark{1},
 Matthieu Perrot\IEEEauthorrefmark{2},
 Clara Fischer\IEEEauthorrefmark{2}, \\
 Vincent Frouin\IEEEauthorrefmark{1}
 and Édouard Duchesnay\IEEEauthorrefmark{1}
 }
 \IEEEauthorblockA{\IEEEauthorrefmark{1}
 Neurospin, I2BM, CEA, Gif-sur-Yvette - France}
 \IEEEauthorblockA{\IEEEauthorrefmark{2}
 Centre d'Acquisition et de Traitement des Images (CATI), Gif-sur-Yvette - France}
 \IEEEauthorblockA{Corresponding author: \texttt{edouard.duchesnay@cea.fr}}
}

\IEEEoverridecommandlockouts
\IEEEpubid{\makebox[\columnwidth]{978-1-4799-4149-0/14/\$31.00 ©2014 IEEE \hfill} \hspace{\columnsep}\makebox[\columnwidth]{ }}
\begin{document}
\bstctlcite{IEEEexample:BSTcontrol} 

\maketitle

\begin{abstract}
The use of machine-learning in neuroimaging offers new perspectives in early diagnosis and prognosis of brain diseases.
Although such multivariate methods can capture complex relationships in the data, traditional approaches provide irregular ($\ell_2$ penalty) or scattered ($\ell_1$ penalty) predictive pattern with a very limited relevance.
A penalty like Total Variation (TV) that exploits the natural 3D structure of the images can increase the spatial coherence of the weight map.
However, TV penalization leads to non-smooth optimization problems that are hard to minimize.
%
We propose an optimization framework that minimizes any combination of $\ell_1$, $\ell_2$, and $TV$ penalties
while preserving the exact $\ell_1$ penalty.
This algorithm uses Nesterov's smoothing technique to approximate the $\TV$ penalty with a smooth function such that the loss and the penalties are minimized with an exact accelerated proximal gradient algorithm.
We propose an original continuation algorithm that uses successively smaller values of the smoothing parameter to reach a prescribed precision while achieving the best possible convergence rate.
This algorithm can be used with other losses or penalties.
The algorithm is applied on a classification problem on the ADNI dataset.
We observe that the $\TV$ penalty does not necessarily improve the prediction but provides a major breakthrough in terms of support recovery of the predictive brain regions.
\end{abstract}

\section{Introduction}


Multivariate machine-learning applied in neuroimaging offers new perspectives in early diagnosis and prognosis of brain diseases.
However, it is essential that the method provides meaningful predictive patterns in order to reveal the neuroimaging biomarkers of the pathologies.
Penalized linear models (such as linear SVM, penalized logistic regression) are often used in neuroimaging since the weight map might provide clues about biomarkers.

In particular, we are interested in penalized logistic regression in order to predict the clinical status of patients from neuroimaging data and link this prediction to known neuroanatomical structures.
When using the $\ell_2$ penalty with such data, the weight maps are dense and potentially irregular (i.e. with abrupt, high-frequency changes).
With the $\ell_1$ penalty, they are scattered and sparse with only a few voxels with non-zero weight.
\todo[inline]{This behaviour of the $\ell_1$ penalty depends entirely on the parameter used. Should this be rephrased?}
In both cases, the weight maps are hard to interpret in terms of neuroanatomy.
The combination of both penalties in Elastic Net (see~\cite{Friedman2010}), promotes sparse models while still maintaining the regularization properties of the $\ell_2$ penalty.
A major limitation of the Elastic Net penalty is that it does not take into account the spatial structure of brain images, which leads to scattered patterns.

%
%
The Total Variation ($\TV$) penalty is widely used in 2D or 3D image processing to account for this spatial structure.
In this paper, we propose to add $\TV$ to the Elastic Net penalty to improve the interpretability and the accuracy of logistic regression.
We hypothesize that the predictive information is most likely organized in regions rather than scattered across the brain.

The difficulty is that $\ell_1$ and $\TV$ are convex but not smooth functions (see~\autoref{sec:method} for the precise definition of smoothness used in this paper).
Therefore, we cannot use classic gradient descent algorithms.
In~\cite{Gramfort2013}, the authors use a primal-dual approach for $\ell_1$ and $\TV$ penalties (which can be extended to include $\ell_2$) but their method is not applicable to logistic regression
because the proximal operator of the logistic loss is not known.
Another strategy for non-smooth problems is to use methods based on the proximal operator of the penalties.
For the $\ell_1$ penalty alone, the proximal operator is analytically known and efficient iterative algorithms such as ISTA and FISTA are available (see~\cite{Beck2009}).
However, as the proximal operator of the $\TV$ penalty is not analytically defined, those algorithms won't work in our case.

There are two general strategies to address this problem.
The first one involves using an iterative algorithm to numerically approximate the proximal operator of each convex non-smooth penalty (see~\cite{Schmidt2011}).
This algorithm is then run for each iteration of ISTA or FISTA (leading to nested optimization loops).
This was done for $\TV$ alone in~\cite{Michel2011} where the authors use FISTA to approximate the proximal operator of $\TV$.
The problem with such methods is that by approximating the proximal operator we may loose the sparsity induced by the $\ell_1$ penalty.
The second strategy is to approximate the non-smooth penalties for which the proximal operator is not known (e.g. $\TV$)
with a smooth function (of which the gradient is known).
Non-smooth penalties with a known proximal operator (e.g. $\ell_1$) are not changed.
Therefore it is possible to use an exact accelerated proximal gradient algorithm.
Such a smoothing technique has been proposed by Nesterov in~\cite{Nesterov2005a}.

We choose to apply the second strategy.
We will present an algorithm able to solve $\TV$-Elastic Net penalized logistic regression with exact $\ell_1$ penalty and evaluate it on the prediction of the clinical status of patients from structural magnetic resonance imaging (MRI) scans.
The paper is organized as follows: we present the minimization problem and our algorithm in~\autoref{sec:method},
the experimental dataset is described in~\autoref{sec:dataset}, and~\autoref{sec:results} presents the classification rates and weight maps.
Finally, we conclude in~\autoref{sec:conclusion}.

\section{Method} \label{sec:method}

We first detail the notations of the problem. Then we develop the TV regularization framework.
Finally, we detail the algorithm used to solve the minimization problem.

\subsection{Problem statement}

We place ourselves in the context of logistic regression models.
Let $X \in \mathbb{R}^{\nSamples \times \nDim}$ be a matrix of $\nSamples$ samples, where each sample lies in a $\nDim$-dimensional space and let $y \in {\{0,1\}}^\nSamples$ denote the $\nSamples$-dimensional response vector.
In the logistic regression model the conditional probability of $y_i$ given the data $X_i$ is defined through
a linear function of the unknown predictors $\beta \in \mathbb{R}^\nDim$ by 
$$
    p_i:=p(y_i=1|X_i) = \frac{1}{1+\exp(- X_i^T \beta)},
$$
and $p(y_i=0|X_i) = 1-p_i$.
Therefore, looking for the maximum of the log-likelihood with structured and sparse penalties, we consider the following minimization problem of a logistic regression objective function with Elastic Net and TV penalties:
\begin{equation} \label{eq:problem}
 \beta^* := \arg \min_{\beta \in \mathbb{R}^\nDim} f(\beta),
\end{equation}
where $f(\beta)$ is the sum of a smooth part, $g(\beta)$, and of a non-smooth part, $h(\beta)$, such that
\begin{eqnarray}\label{Logistic} \nonumber \label{eq:f}
 f(\beta) & := & \underbrace{\frac{1}{\nSamples}\sum_{i=1}^\nSamples \left\{ y_i X_i \beta - \log\left[ 1 + \exp(X_i\beta) \right] \right\}+ \LMltwo \left\|\beta\right\|_2^2}_{g(\beta)} \\
          &  + & \underbrace{\LMlone \left\|\beta\right\|_1 + \LMTV TV(\beta)}_{h(\beta)},
\end{eqnarray}
where $\LMltwo$, $\LMTV$ and $\LMlone$ are constants that control the relative strength of each penalty.
In this context, a function is said to be smooth if it is differentiable everywhere and its gradient is Lipschitz-continuous.

Given a 3D image $I$ of size $(\nDim_x, \nDim_y, \nDim_z)$, $\TV$ is defined as
\begin{equation} \label{eq:TV}
        \TV(I) = \sum_{(i,j,k)} \left\| \grad_{i,j,k}(I) \right\|_2
\end{equation}
where $\grad_{i,j,k}(I) \in \mathbb{R}^3$ is the numerical gradient of $I$ at coordinates $(i,j,k)$
and the sum runs over all voxels of $I$.

In our case, rows of $X$ are composed of masked and flattened 3D images arranged into vectors of size $\nDim<\nDim_x \times \nDim_y \times \nDim_z$.
Similarly, the vector $\beta$ belongs to ${\mathbb{R}}^\nDim$.
For now on each voxel is identified by its linear index in $X$, noted $i$ ($1 \leq i \leq p$).
Special care must be taken for the computation of the gradient on the flattened vector $\beta$, because,
due to the existence of a mask and border conditions, not all the neighbors of a voxel $i$ exist in the data.
Given this precaution, we can compute the gradient for each $\beta_i$ and then compute $\TV(\beta)$.
More details regarding the TV penalty in the context of 3D image analysis can be found in~\cite{Michel2011}.

\subsection{Regularization framework}

A sufficient condition for the application of Nesterov's smoothing technique to a given convex function \(s\) is that it can be written on the form
\begin{equation} \label{eq:dual_constraint}
    s(\beta) = \max_{\boldsymbol{\alpha} \in K_s} \langle\boldsymbol{\alpha}\,|\,\A_s\beta\rangle,
\end{equation}
for all \(\beta \in \mathbb{R}^\nDim\), with \(K\) a compact convex set in a finite-dimensional vector space and
\(\A_s\) a linear operator between two finite-dimensional vector spaces.

In~\cite{Lofstedt2014} the authors show that $\TV(\beta)$ can be written as
\begin{equation*}
    \TV(\beta) = \sum_{i=1}^{p} \max_{\alpha_{i} \in K_{i}} \langle \alpha_{i} | A_{i} \beta \rangle
\end{equation*}
where $K_{i} = \left\{ \alpha \in \mathbb{R}^3, \|\alpha\|_2^2 \leq 1 \right\}$
and $A_{i}$ is a sparse matrix that allows to compute the gradient at position $i$ ($A_{i}$ depends on the mask $\mask$).
This can be further written as
\begin{equation*}
    \TV(\beta) = \max_{\alpha \in K} \langle \alpha | \A \beta \rangle
\end{equation*}
where $\alpha$ is the concatenation of all the $\alpha_{i}$,
$\A$ is the vertical concatenation of all the $A_{i}$ matrices and
$K$ is the product of all the compact convex spaces $K_{i}$ (as such, $K$ is itself a compact convex space).
Note that $K$ and $\A$ are specific to $\TV$.

Given this expression for $\TV$, we can apply Nesterov's smoothing.
For a given smoothing parameter $\mu > 0$, $\TV$ is approximated by the smooth function
\begin{equation} \label{eq:dual_constraint_smoothed}
    \approxTV{\mu}(\beta) = \max_{\alpha \in K} \left\lbrace \langle\alpha|\A\beta\rangle - \frac{\mu}{2}\|\alpha\|_2^2 \right\rbrace.
\end{equation}
The value that maximizes~\autoref{eq:dual_constraint_smoothed} is
\begin{equation*}
    \alphaApprxNest{\mu}(\beta) = \proj_{K} \left( \frac{A\beta}{\mu} \right)
\end{equation*}
The function \(\approxTV{\mu}\) is convex and differentiable.
Its gradient can be written (see~\cite{Nesterov2005a}) as
\begin{equation*}
    \nabla\approxTV{\mu}(\beta) = \A^\T\alphaApprxNest{\mu}(\beta).
\end{equation*}
The gradient is Lipschitz continuous with Lipschitz constant
\[
    \frac{\|\A\|_2^2}{\mu},
\]
where \(\|\A\|_2\) is the matrix spectral norm of \(\A\).

\subsection{Algorithm}

A new optimization problem, closely related to problem~\ref{eq:problem}, arises from this regularization:
\begin{equation} \label{eq:problem_mu}
    \optSmoothedBeta := \arg \min_{\beta \in \mathbb{R}^\nDim} f_{\mu}(\beta)
\end{equation}
where
\begin{equation} \label{eq:f_mu}
    f_{\mu}(\beta) := \underbrace{g(\beta) + \LMTV TV_{\mu}(\beta)}_{\text{smooth}} + \underbrace{\LMlone \left\|\beta\right\|_1}_{\text{non-smooth}}.
\end{equation}
$\optSmoothedBeta$ approximates $\beta^*$, the solution to the original problem~\ref{eq:problem}, since $\left\| f_\mu - f \right\| \leq \frac{\mu \nDim}{2}$.

Since we are now able to explicitly calculate the gradient of the smooth part, its Lipschitz constant and the proximal operator of the non-smooth part, this new problem can be solved by FISTA~\cite{Beck2009}.
The convergence rate of FISTA is governed by
\begin{equation}\label{eq:FISTA_convergence_upper_bound}
    f_{\mu}(\beta^{(k)}) - f_{\mu}(\optSmoothedBeta) \leq \frac{2}{t_\mu(k+1)^2}\|\beta^{(0)} - \optSmoothedBeta\|_2^2,
\end{equation}
where $k \geq 1$ is the iteration number and $t_\mu$ is the step size that must be chosen smaller than or equal to the inverse of the known Lipschitz constant of the gradient of the smooth part.
Note that the convergence depends on the initial value $\beta^{(0)}$.

If $\mu$ is small the algorithm will converge with a high precision (i.e. $\optSmoothedBeta$ will be close to $\beta^*$) but
in this case it will converge slowly (because small $\mu$ leads to small $t_{\mu}$).
Thus, there is a trade-off between speed and accuracy.
We therefore propose to perform successive runs of FISTA with decreasing values of the smoothing parameter (to increase precision)
but using the regression vector obtained at the previous run as a starting point for FISTA to increase convergence speed.
We denote $\beta^{(i)}$ the regression vector after the $i$th run of FISTA.

The key point is how to derive the sequence of smoothing parameter $\mu^{(i)}$. 
Our approach involves two steps.
First, we describe how to obtain a value of the smoothing parameter $\mu_{opt}(\varepsilon)$ that minimizes the number of iterations needed to achieve a prescribed precision $\varepsilon > 0$ when minimising \ref{eq:problem} via \ref{eq:problem_mu} (i.e. such that $f(\beta^{(k)}) - f(\beta^*) < \varepsilon$).
Next, given a predefined sequence $\varepsilon^{(i)}$ of decreasing precision values, we can define a continuation sequence of smoothing parameters such that $\mu^{(i)} = \mu_{opt}(\varepsilon^{(i)})$.
Concerning the first point we can prove that for any given $\varepsilon>0$, selecting the smoothing parameter as
\begin{equation*} \label{eq:mu_opt}
    \mu_{opt}(\varepsilon) = \frac{-\LMTV \|A\|_2^2}{L_0} + \frac{\sqrt{(\LMTV M \|A\|_2^2)^2 + \varepsilon M L_0\|A\|_2^2}}{M L_0}
\end{equation*}
where $M=\nDim/{2}$ and $L_0$ is the Lipschitz constant of $\nabla(g)$ (following~\cite{Michel2011}, we have $L_0=2\LMltwo+\|A\|_2/(4\nSamples)$)
minimizes the worst case bound on the number of iterations needed to achieve the precision $\varepsilon$ when minimizing \ref{eq:problem} via \ref{eq:problem_mu}.
The proof is inspired by the proof of Lemma~3 in~\cite{Savchynskyy2011}.
In this article, we use a fixed sequence of precision $\varepsilon^{(i)} = (1/2)^{i-1}$.
The only parameter of the algorithm is then the initial point $\beta^0$.
In these experiments, we used a random vector with a unit norm.


We call this algorithm CONESTA (for \textbf{CO}ntinuation with \textbf{NE}sterov smoothing in a \textbf{S}hrinkage-\textbf{T}hresholding \textbf{A}lgorithm).
The algorithm is presented in~\autoref{alg:CONESTA}.
The convergence proof will be presented in an upcoming paper.
We denote the total number of FISTA loops used in CONESTA by $K$.
We have experimentally verified that the convergence rate to the solution of problem~\ref{eq:problem} is $\bigO{\nicefrac{1}{K^2}}$ (which is the optimal convergence rate).
Also, the algorithm works even if some of the weights $\LMlone$, $\LMltwo$ or
$\LMTV$ are zero, which thus allows us to solve e.g. the Elastic Net or pure lasso using CONESTA.

\begin{algorithm}[!htp]
    \caption{CONESTA}
    \label{alg:CONESTA}
    \begin{algorithmic}[1]
        \REQUIRE $\beta^{0}$, the initial regression vector.
        \STATE $i = 1$
        \REPEAT
         \STATE $\epsilon^{(i)} \leftarrow (1/2)^{i-1}$
         \STATE $\mu^{(i)} \leftarrow \mu_{opt}(\epsilon^{(i)})$
         \STATE $\beta^{(i)} \leftarrow FISTA(\beta^{(i-1)}, \mu_i)$
         \STATE $i = i + 1$
        \UNTIL{Convergence}
    \end{algorithmic}
\end{algorithm}



\section{Dataset} \label{sec:dataset}

The data used in the preparation of this article were obtained from the Alzheimer's Disease Neuroimaging Initiative (ADNI) database (\url{http://adni.loni.usc.edu/}).
The MR scans are T1-weighted MR image acquired at 1.5~T according to the ADNI acquisition protocol (see~\cite{Jack2008}).
The image dimensions were $\nDim_x=\pxval$, $\nDim_y=\pyval$, $\nDim_z=\pzval$.
The \nImagesADNI{} T1-weighted MR images were segmented into GM (Gray Matter), WM (White Matter) and CSF (Cerebrospinal Fluid) using the SPM8 unified segmentation routine~\cite{Ashburner2005}.
\numprint{\nImagesQC} images were retained after quality control on GM
probability. These images were spatially normalized using DARTEL~\cite{Ashburner2007} without any spatial smoothing.
From the \numprint{\nImagesQC} registered images we use only the \numprint{\nControl} control (CTL) subjects and the \numprint{\nAD} Alzheimer's Disease (AD) subjects.
Thus, the total number of images was $n=\numprint{\nImages}$.
A brain mask was obtained by thresholding the modulated gray matter map, leading to the selection of $\nDim=\numprint{\pval}$ voxels.
According to the assignments found in~\cite{Cuingnet2011}, those \numprint{\nImages} images were split into $\nTrainingImages$ training images, used in the learning phase, and $\nTestingImages$ images used to test the algorithms.


\section{Experimental results} \label{sec:results}


\autoref{tab:results} presents the prediction results obtained on the test samples.
It shows that using the $\ell_1$ penalty alone decreases the predictive performance.
We suspect that the $\ell_1$ penalty is inefficient in recovering the predictive support on non-smoothed images.
The $\TV$ penalty does not significantly increase nor decrease the performances except when it is combined with the $\ell_1$ penalty.

\begin{figure}
 \centering
 \includegraphics[width=\textwidth/2]{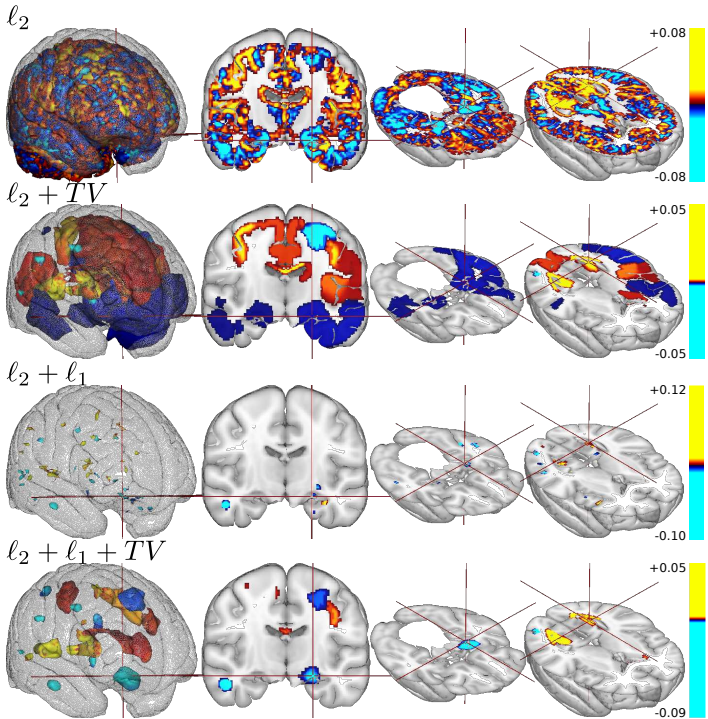}
 \caption{Weight maps: positive/negative values indicate the way regions contribute to predict the AD status.
          It should generally be interpreted as a increase/decrease of GM in the AD group.
          Positive weights (increase of GM in AD) may be found where negative weights are expected.
          For example, positive weights surround the whole bi-lateral ventricles.
          We hypothesize that we observe the negative pattern of an underlying global atrophy: the GM surrounding the ventricles shift away from them thus we observe GM in AD patients where controls have WM tissue.
          The map obtained with $\ell_1+TV$ has been omitted since it provides similar results as those found with $\ell_1+\ell_2+TV$.
          The map obtained with $\ell_1$ alone has no relevance.}
 \label{fig:betamaps}
\end{figure}

\autoref{fig:betamaps} demonstrates that the $TV$ penalty provides a major breakthrough in terms of support recovery of the predictive brain regions.
Conversely to the $\ell_2$ penalty that highlights an irregular and meaningless pattern, $\ell_2 + TV$ provides a smooth map that match the well-known brain regions involved in AD \cite{Frisoni2010}.
A large region of negative weights was found in the temporal lobe.
This region includes the superior and middle temporal gyri, the parahippocampal gyrus and the entorhinal cortex, the fusiform gyrus, the amygdala, the insula and the hippocampus.
As expected, this pattern was predominantly found on the left hemisphere.
The bi-lateral ventricular enlargement is sharply identified, the surprising positive sign of the weights is explained in \autoref{fig:betamaps}.
Atrophy in the frontal lobe (inferior frontal gyrus) was found. 
Positive weights within the whole cingulum region reflect tissue shift due to peri-ventricular atrophy. 
In the occipital lobe, positive weights were observed within the calcarine fissure and the cuneus.

As hypothesized, the combination of the $\ell_1+\ell_2$ penalties provides scattered patterns with a very limited relevance.

Finally, $\ell_1+\ell_2+TV$ provides a summary of the $\ell_2+TV$ pattern: most of the identified regions are are the same as when using $\ell_2+TV$ but with limited extent.
For example, the whole temporal atrophy found by $\ell_2+TV$ is now limited to the hippocampus.
Noticeably, the right hippocampus is no longer a predictive region due to the property of the $\ell_1$ penalty.
This suggests that sparse patterns should be considered with caution.

\begin{table}
 \centering
 \caption{Prediction accuracies. Sensitivity (Sens.: recall rate of AD patients), Specificity (Spec.: recall rate of CTL subjects), BCR (Balanced Classification Rate) 
 and McNemar's comparison test $p$-value against another method. All prediction rates were significant except those obtained with the $\ell_1$ method.}
\begin{tabular}{llcccc}
\toprule
Method             & $\LMltwo, \LMlone, \LMTV$ & Sens. & Spec. & BCR   & Comp. $p$-value\\
\midrule
$\ell_2$           & $1.0, 0.0, 0.0$           &  0.855      &  0.855      & 0.855 & -\\
$\ell_1$           & $0.0, 1.0, 0.0$           &  0.684      &  0.484      & 0.584 & -\\
$\ell_2+\ell_1$    & $0.9, 0.1, 0.0$           &  0.802      &  0.742      & 0.772 & -\\
\midrule
$\ell_2+TV$        & $0.1, 0.0, 0.9$           &  0.842      &  0.726      & 0.784 & 0.16 to $\ell_2$\\
$\ell_1+TV$        & $0.0, 0.1, 0.9$           &  0.829      &  0.774      & 0.801 & \textbf{2e-4} to $\ell_1$\\
$\ell_2+\ell_1+TV$ & $0.1, 0.1, 0.8$           &  0.815      &  0.758      & 0.787 & 1 to $\ell_2+\ell_1$\\
\bottomrule
\end{tabular}
 \label{tab:results}
\end{table}

\section{Conclusion} \label{sec:conclusion}

We proposed an optimization algorithm that is able to minimize any combination of the $\ell_1$, $\ell_2$, and $TV$ penalties
while preserving the exact $\ell_1$ penalty.
This algorithm uses Nesterov's technique to smooth the $TV$ penalty such that objective function is minimized with an exact accelerated proximal gradient algorithm.
The approximation of $\TV$ is controlled by a single smoothing parameter $\mu$ .
Our contribution was to propose a continuation algorithm with successively smaller values of $\mu$ to reach a prescribed precision while achieving the best possible convergence rate.
Average execution time is one hour on a standard workstation involving \numprint{13000} FISTA iterations.

We observed that by adding the $\TV$ penalty, the prediction does not necessarily improve.
However, we demonstrated that it provides a major breakthrough in terms of support recovery of the predictive brain regions.

It should be noted that the algorithm can be extended to minimize any
differentiable loss (logistic, least square) with any combination of $\ell_1$,
$\ell_2$ penalties and with any non-smooth penalty that can be written in the form of~\autoref{eq:dual_constraint}.
This includes Group Lasso and Fused Lasso or any penalty that can be expressed as a $p$-norm of a linear operation on the weight map.

\section*{Acknowledgement}
This work was partially funded by grants from the French National Research Agency ANR BRAINOMICS (ANR-10-BINF-04) and from the European Commission MESCOG (FP6 ERA-NET NEURON 01 EW1207).

\bibliographystyle{IEEEtran}
\bibliography{IEEEabrv,bib_control,prni_2014}

\end{document}